\documentclass[letterpaper, 10 pt, conference]{ieeeconf}  

\IEEEoverridecommandlockouts                              

\usepackage{booktabs} 		
\usepackage{makecell}
\usepackage{colortbl}

\usepackage{algcompatible}
\usepackage{amsfonts}
\usepackage{multirow}
\usepackage{makecell}
\usepackage{booktabs}
\usepackage{xcolor}
\usepackage{balance}
\definecolor{trainrefined}{RGB}{224, 88, 61}
\definecolor{trainsmalltalk}{RGB}{107, 69, 209}
\definecolor{test}{RGB}{54, 133, 101}
\definecolor{val}{RGB}{133, 122, 45}

\newcommand{\mypar}[1]{\vspace{0.3cm}\noindent\textbf{#1}}
\usepackage{graphicx}
\usepackage{lipsum}
\newcommand{\relu}[1]{\ensuremath{\text{ReLU}}}
\usepackage{subcaption}
\usepackage{pifont}

\usepackage{tikz,xcolor,hyperref}
\definecolor{lime}{HTML}{A6CE39}
\DeclareRobustCommand{\orcidicon}{
	\begin{tikzpicture}
	\draw[lime, fill=lime] (0,0) 
	circle [radius=0.16] 
	node[white] {{\fontfamily{qag}\selectfont \tiny ID}};
	\draw[white, fill=white] (-0.0625,0.095) 
	circle [radius=0.007];
	\end{tikzpicture}
	\hspace{-2mm}
}
\foreach \x in {A, ..., Z}{\expandafter\xdef\csname orcid\x\endcsname{\noexpand\href{https://orcid.org/\csname orcidauthor\x\endcsname}
			{\noexpand\orcidicon}}
}

\title{\LARGE \bf
On Transferability of  Driver Observation Models  from \\ Simulated to Real Environments in  Autonomous Cars
}

\author{Walter Morales-Alvarez$^{1*}$\orcidW{} \emph{Member, IEEE}, Novel Certad$^{1*}$\orcidN{} \emph{Graduate Student Member, IEEE}, \\Alina Roitberg$^{2*}$ \orcidA{}  \emph{Member, IEEE}, Rainer Stiefelhagen$^{3}$ \orcidR{}  \emph{Member, IEEE} and \\Cristina Olaverri-Monreal$^{1}$\orcidC{} \emph{Senior Member, IEEE}%
\thanks{*Denotes equal contribution.}
\thanks{$^{1}$Chair ITS-Sustainable Transport Logistics 4.0, Johannes Kepler University Linz, Altenberger Straße 69, 4040 Linz, Austria.
\texttt{\{novel.certad\_hernandez, walter.morales\_alvarez, cristina.olaverri-monreal\}@jku.at}}
\thanks{$^{2}$Institute for AI, University of Stuttgart, 70569 Stuttgart, Germany.
\texttt{alina.roitberg,@f05.uni-stuttgart.de}}%
\thanks{$^{3}$Institute for Anthropomatics and Robotics, Karlsruhe Institute of Technology,
76131 Karlsruhe, Germany.
\texttt{rainer.stiefelhagen@kit.edu}}%
}

\begin{document}

\maketitle
\thispagestyle{empty}
\pagestyle{empty}


\begin{abstract}

For driver observation frameworks, clean datasets collected in controlled simulated environments often serve as the initial training ground. Yet, when deployed under real driving conditions, such simulator-trained models quickly face the problem of distributional shifts brought about by changing illumination, car model, variations in subject appearances, sensor discrepancies, and other environmental alterations.


This paper investigates the viability of transferring video-based driver observation models from simulation to real-world scenarios in autonomous vehicles, given the frequent use of simulation data in this domain due to safety issues. To achieve this, we record a dataset featuring  actual autonomous driving conditions and involving seven participants engaged in highly distracting secondary activities.
To enable   direct \textsc{Sim$\rightarrow$Real} transfer, our dataset was designed in accordance with an existing  large-scale simulator dataset used as the training  source. 
We utilize the Inflated 3D ConvNet (I3D) model, a popular choice for driver observation, with Gradient-weighted Class Activation Mapping (Grad-CAM) for detailed analysis of model decision-making.
Though the simulator-based model clearly surpasses the random baseline, its recognition quality  diminishes, with average accuracy dropping from 85.7\% to 46.6\%. We also observe strong variations across different  behavior classes. This underscores the challenges of model transferability, facilitating our research of 
more robust  driver observation systems capable of dealing with real driving conditions.

\end{abstract}



\section{Introduction}

To speed up the development of driver observation systems, researchers often leverage simulated environments for collecting the training data~\cite{martin2019drive,tran2020realtime_detection_distracted,katrolia2021ticam,kopuklu2021driver}. These environments provide a controlled and easily reproducible setting, which allows for the collection of clean datasets, avoiding the challenges associated with real-world complexities.
 Furthermore, when studying driver behaviors during autonomous or highly automated driving~\cite{martin2019drive,morales2022automated,gold2013take,mok2015emergency,flad2020personalisation}, safety  becomes a significant challenge, leading to simulators being very prominent.
However, deploying models trained in simulation to real-world scenarios presents a significant challenge, as the distributional shifts between the two environments can lead to poor generalization and performance degradation. 
These shifts can be caused by various factors, such as changing lighting conditions, discrepancies in car models and sensors, and other environmental variations. 
Some works have presented results  from real world tests to validate and show potential discrepancies regarding results from simulated data, e.g., for take over requests~\cite{morales2020automated}.

\begin{figure}
    \centering
    \includegraphics[width=0.5\textwidth]{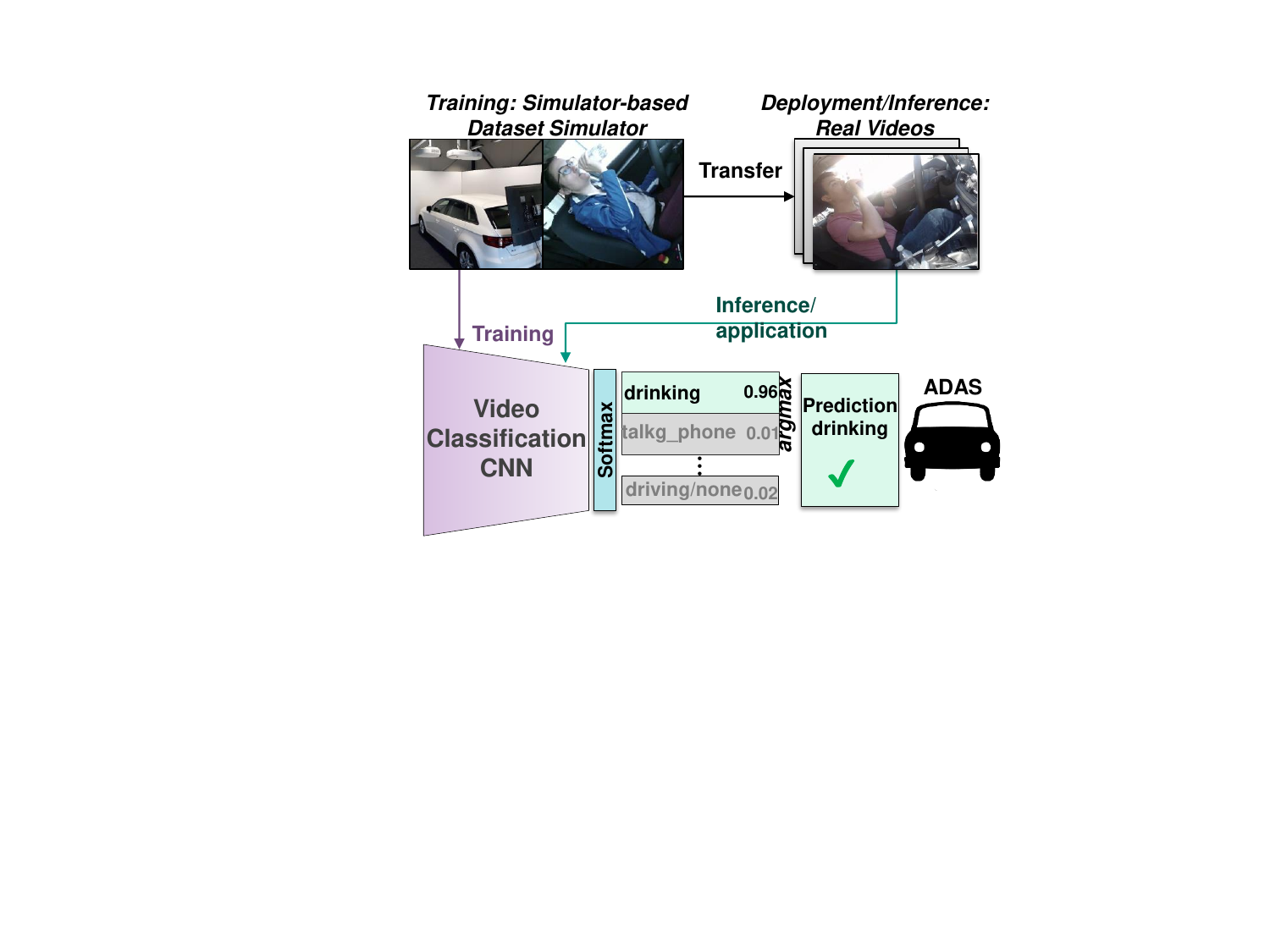}
    \caption{We collect the first video-based driver activity recognition dataset featuring  secondary activities in \textit{real autonomous driving} scenarios. Unlike the controlled simulated environments, our recordings include real-world complexities, such as car movement and fluctuating lighting conditions.  Combined with a large-scale training database collected in a simulator, we utilize our dataset as a testbed for validating direct \textsc{Sim$\rightarrow$Real} transfer of deep learning-based driver observation models.
}
    \label{fig:teaser}
\end{figure}

To address these issues and improve the transferability of driver observation models, this paper investigates the efficacy of different CNN-based approaches in bridging the domain gap between simulated and real environments (an overview is provided in Figure ~\ref{fig:teaser}).
We first introduce the validation testbed, a collected dataset for video-based driver monitoring in autonomous vehicles, and the training dataset generated from a simulated environment.  Through a series of experiments, we provide a thorough evaluation of the performance of these models in real driving scenarios.

By offering valuable insights into the challenges associated with transferring models from simulation to real-world scenarios, we aim to contribute to the development of more robust and reliable driver observation systems that can be deployed in real driving conditions. 

\begin{table}[b!]
  \centering
  \caption{Overview of the recorded video  dataset of secondary activities during real autonomous driving sessions.}
  \label{tab:data_summary}
  \begin{tabular}{ l c }
 
      \specialrule{1pt}{0pt}{0pt}
    \rowcolor{gray!20}\multicolumn{2}{c}{\textbf{Dataset statistics}} \\
    \specialrule{1pt}{0pt}{0pt}
    Context & Real driving session \\
    Manual driving & $\checkmark$ \\
    Autonomous driving & $\checkmark$ \\
    Data type & RGB video \\
    Number of subjects & 7 \\
    Nr. female subjects & 2 \\
    Nr. driver activities & 7 \\
    Recording lengths (min) & 100.83 \\
    Number of samples* & 1987 \\
    \specialrule{1pt}{0pt}{0pt}
  \end{tabular}
\end{table}

\begin{figure}[b!]
    \centering
    \includegraphics[width=0.3\textwidth]{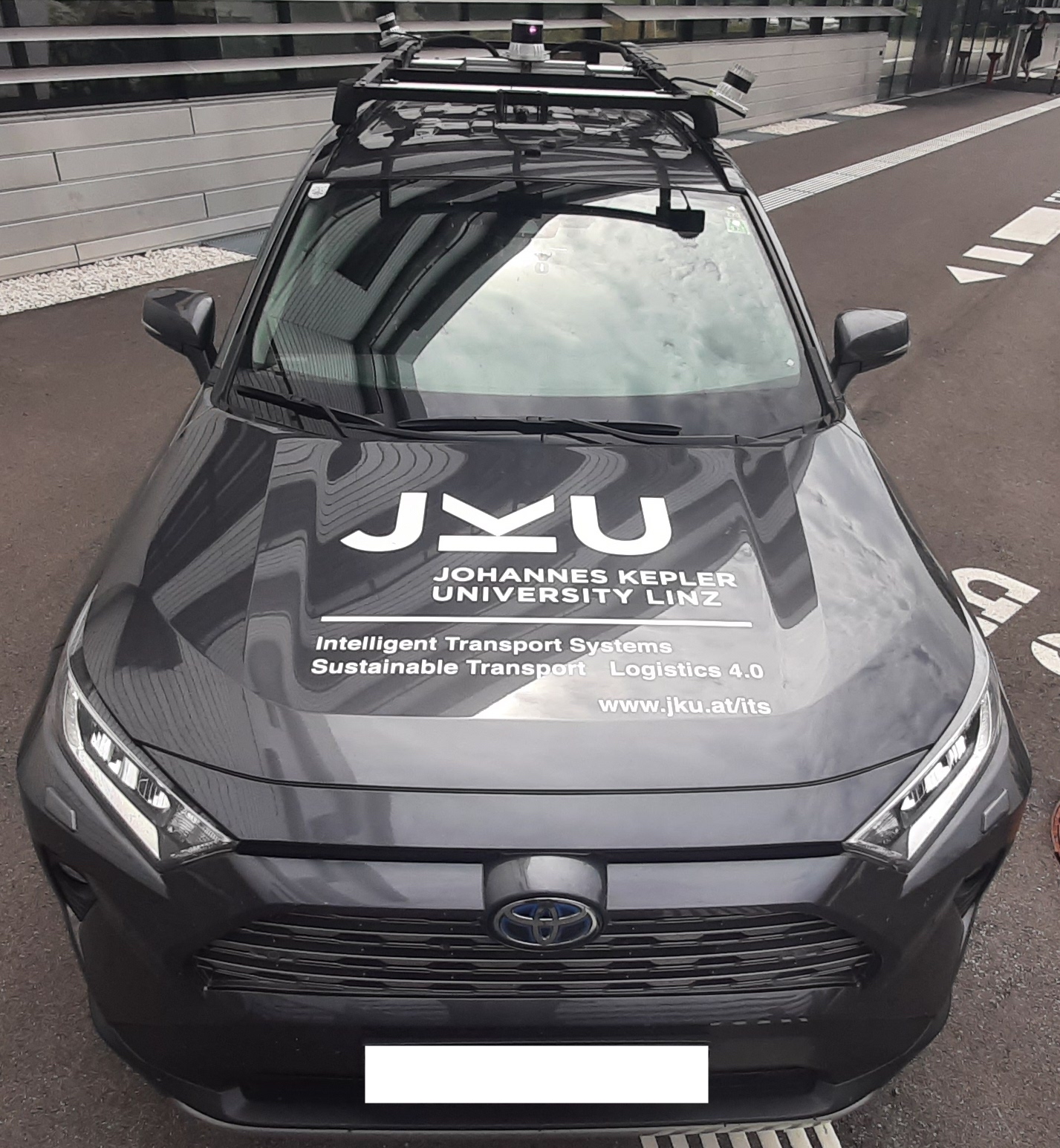}
    \caption{Autonomous vehicle used for the data collection.}
    \label{fig:car}
\end{figure}


\section{Related Work}

Models for recognizing driver activity from video fall into two main categories: ones that utilize manual feature descriptors, and ones that leverages end-to-end deep learning, processing video inputs directly and concurrently learning intermediate representations alongside the classifier.
Traditional methodologies based on manual features~\cite{ohn2014head_eye_hand,xu2014realtime_random_forests,zheng2015eye_gaze,braunagel2015driver_conditionally,ohn2014head_eye_hand,braunagel2015driver_conditionally} exploit classical machine learning models like Support Vector Machines and Random Forest. These first extract  features related to the driver's hand movements, body and head posture, and eye direction.
Recently, end-to-end deep learning models have become an increasingly prevalent choice for recognizing driver activities. These models often utilize Convolutional Neural Networks (CNNs)\cite{tran2020realtime_detection_distracted, martin2019drive, tan2021bidirectional,roitberg2022my,peng2022transdarc,roitberg2020open} and transformer-based models\cite{peng2022transdarc} as their backbone.
While neural network-based approaches are at the top of most driver observation benchmarks, they require a significant amount of annotated training data. Several real-world datasets are tailored for in-vehicle observation during \textit{manual} driving~\cite{jain2016recurrent,xing2019driver,ohn2014head_eye_hand}.
However, for \textit{autonomous} driving, simulators are more commonly used~\cite{martin2019drive,morales2022automated,gold2013take,mok2015emergency}.
For instance, Drive\&Act~\cite{martin2019drive}, the largest public dataset for video-based driver observation during autonomous driving, was collected in a stationary car placed indoors, surrounded by three screens imitating outdoor surroundings.
Conversely, research from the broader field of human activity recognition reports a substantial decline in recognition quality when transitioning from synthetic to real data~\cite{roitberg2021let}.
Another line of work focuses on domain adaptation for driver observation, such as cross-view recognition or model adaptation for participants wearing glasses~\cite{reiss2020deep, rangesh2020driver}. Given real-life training data and resources for post-hoc model adjustment, these domain adaptation approaches show promising potential to significantly enhance recognition, presenting an important future research direction. However, this falls outside the scope of our study, as such methods require additional unlabeled training data in the target domain and cannot be applied directly.
Overall, recent research in video-based driver observation tends to prioritize the development of high-accuracy classifiers for conditions similar to training environments, while performance under distributional shifts or adverse conditions is often considered secondary.

Inspired by this, we present an empirical evaluation of direct \textsc{Sim$\rightarrow$Real} transfer of deep learning-based activity recognition models in the context of autonomous driving.
To address the gap in suitable \textsc{Sim$\rightarrow$Real} benchmarks, we first collect a real-world dataset of in-vehicle observation during actual autonomous driving, annotated with a subset of activities present in a large-scale simulator-based dataset, facilitating the aforementioned validation scenario.

\section{Validating  Simulator-based Driver Observation Models in Real  Environments}

\subsection{Testbed: collected dataset for video-based driver monitoring in autonomous vehicles}

The dataset used in this study consisted of videos obtained from the scientific personnel of Johannes Kepler University Linz. The data collection was carried out utilizing the JKU-ITS vehicle (see Figure \ref{fig:car}), as described in \cite{certad2023jkuits}, where participants engaged in various tasks while the vehicle autonomously navigated through a designated test lane within the university premises.

The data collection setup resembled that of the work presented in \cite{vallebarrio2023development}, where participants were required to activate the vehicle's automation and commence task performance, while the automation system assumed control of the vehicle throughout the experiment.

In our study, the vehicle automation involved two distinct processes. The first process encompassed the drive-by-wire capability, implemented using Openpilot algorithms \cite{openpilot}. By utilizing the Black Panda device, acceleration and steering commands were transmitted to the vehicle via the internal ADAS (Advanced Driver Assistance System), facilitated by a ROS Wrapper that exposed the Black Panda's communication protocols to ROS topics. The second process involved a custom ROS2 high-level controller of the vehicle, responsible for generating trajectories, speed profiles, and steering and acceleration commands based on pre-recorded waypoints obtained through the vehicle's GPS.

To ensure safety during the experiment, a security driver was present in the passenger seat, overseeing the proper functioning of the system. In case of emergencies, the security driver could assume manual control of the vehicle using a joystick to apply the brakes.

For video recording of the participants, a Logitech C920 webcam was utilized. The webcam was positioned on the passenger door, capturing the entirety of the participants' body movements.

\mypar{Dataset statistics}. 
Table \ref{tab:data_summary} provides an overview of a recorded video dataset of secondary activities during real autonomous driving sessions. All secondary activities are recorded during fully autonomous driving, except for the \textit{driving/sitting still} activity, which also included sequences of the subject steering manually.
The dataset includes seven subjects in total, two female and five male. These subjects are recorded engaging in seven different driver activities: driving/sitting still, using a phone, talking on the phone, reading a magazine, reading a newspaper, reading a book, and drinking. The duration of the recorded data is  $100.83$ minutes.
Following the procedure of~\cite{martin2019drive} a single sample to be classified is defined as a 3-second video clip that is labeled with a specific activity. The objective of the recognition model is therefore to accurately tag each 3-second or shorter action segment (for events of lesser duration) with the appropriate activity label. The dataset comprises $1987$ such annotated samples. Note, that our dataset is not intended for training, but for validation of the \textsc{Sim$\rightarrow$Real} transfer of modern neural networks trained on simulator-based data.

\subsection{Validation Protocol and Recognition Model}

\mypar{Training dataset collected in a simulator.}
As the simulator-based training dataset, we leverage Drive\&Act\cite{martin2019drive}, the largest public in-vehicle human activity dataset focused on distractive behavior during both, manual and autonomous driving. The data is collected from $15$ subjects  and is annotated with $34$ fine-grained activities at the main evaluation level.
To maintain label correspondences, we select $6$ categories present in Drive\&Act. In addition, we collect the \emph{reading book} activity, which was not present in Drive\&Act in this form and is therefore interesting for looking at the networks behavior in the case of a new object (book). At the same time, Drive\&Act contains similar activities - \emph{reading newspaper} and \emph{reading magazine} and an ideal model would map the new \emph{reading book}  situation to one of these states

\mypar{Neural Architecture.}
We utilize the Inflated 3D architecture (I3D)\cite{carreira2017quo}, an extension of the Inception-v1 network, renowned in video classification and driver observation fields\cite{martin2019drive}. The I3D adapts the 2D filters of Inception-v1 into a temporal dimension and processes 64-frame video snippets of 224x224 resolution. I3D consists of 27 layers, with nine Inception modules executing parallel convolutions and concatenating the output to ensure computational efficiency.
We use the original model~\cite{martin2019drive} trained on the Drive\&Act  split 1 (200 epochs, SGD at a learning rate of 0.05 and momentum of 0.9, pre-training on Kinetics) and exclude the activity labels not present in our dataset in the last fully-connected layer.

\subsection{Model Attribution Analysis}
\label{sec:gradcam}

To examine the way simulator-based video classification models operate when facing real-life driving data, we leverage  the Gradient-weighted Class Activation Map method (GradCAM)~\cite{selvaraju2016grad}. 
However,  the original approach was designed for static images~\cite{selvaraju2016grad}, while we are additionally dealing with the time dimension.  We, therefore, implement a three-dimensional variant of GradCAM similar to ~\cite{roitberg2020cnn_spatialtemporal}.

 Given an input video, we predict a class $c$, then estimate the gradient over $y_c$ with respect to each value in the feature channel $A_k$.

The \emph{importance} $w_{c}^{k}$ is then estimated separately for each channel $k$ by averaging the gradients:

\begin{equation}
w_{c}^{k} = \frac{1}{n}\sum_{i, j, t} \Big(\frac{\partial{y_{c}} }{\partial{A_{k}^{i, j, t}}}\Big),
\end{equation}

We then calculate final weights $V_c^{i, j, t}$ by applying the $ReLU$ function to the linear combination of feature map values and importance estimates:

\begin{equation}
V_{c}^{i, j, t} = ReLU \big( \sum_{k}w_{c}^{k}A_{k}^{i, j, t}\big).
\end{equation}

For visualization, we average the heat-maps over time.

\section{Validation Results}
\subsection{Main quantitative results}
We evaluate the recognition quality of driver activities collected in a simulator versus a real-life self-driving car, as presented in Table~\ref{tab:accuracy}. The performance of each model is evaluated based on accuracy.

For all driver activities, a random chance baseline yields an accuracy of 14.29\% in both the simulator and real-life scenarios. The Inflated 3D architecture (I3D), when trained on simulator data, attains an overall accuracy of 85.7\% in the simulator and 46.56\% in real-life observations. Notably, the I3D model significantly outperforms the random baseline in both settings, demonstrating its efficacy.

\begin{figure}
    \centering
    \includegraphics[width=0.45\textwidth]{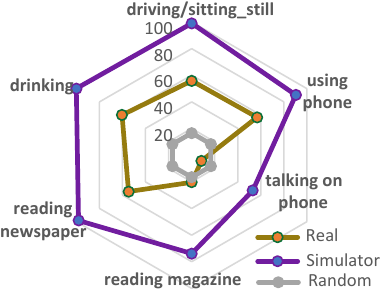}
    \caption{Accuracy of the individual categories visualized  a radar diagram. We compare  the recognition quality of the I3D  model trained on a simulator-based dataset and validated on (1) data originating from the same simulator  (2) in-vehicle observation videos during real autonomous driving sessions collected by us, and (3) random chance baseline.}
    \label{fig:radar}
\end{figure}

However, the per-category results for the simulator-trained I3D reveal some variability. In the simulator, high accuracy is observed for activities such as \textit{driving/sitting still} (99.01\%), \textit{using a phone} (90.36\%), \textit{reading a newspaper} (98.03\%), and \textit{drinking }(100\%). However, the accuracy decreases considerably in a real-life environment, with the highest scores being \textit{reading a book} (69.49\%) and \textit{drinking} (60.4\%). Notably, \textit{talking on the phone} and \textit{reading a magazine} activities witness the most substantial drops in accuracy, scoring merely 8.52\% and 20.34\%, respectively, in real-life conditions. To facilitate the comparison between the different accuracy results, they are graphically represented in Figure \ref{fig:radar} as a radar diagram.

This analysis underscores the difficulties in transferring models trained in simulated environments to real-world conditions, especially for specific activities. It also suggests the need for further fine-tuning and optimization of the I3D model for real-life driver observation.

\begin{figure}
    \centering
    \includegraphics[width=0.5\textwidth]{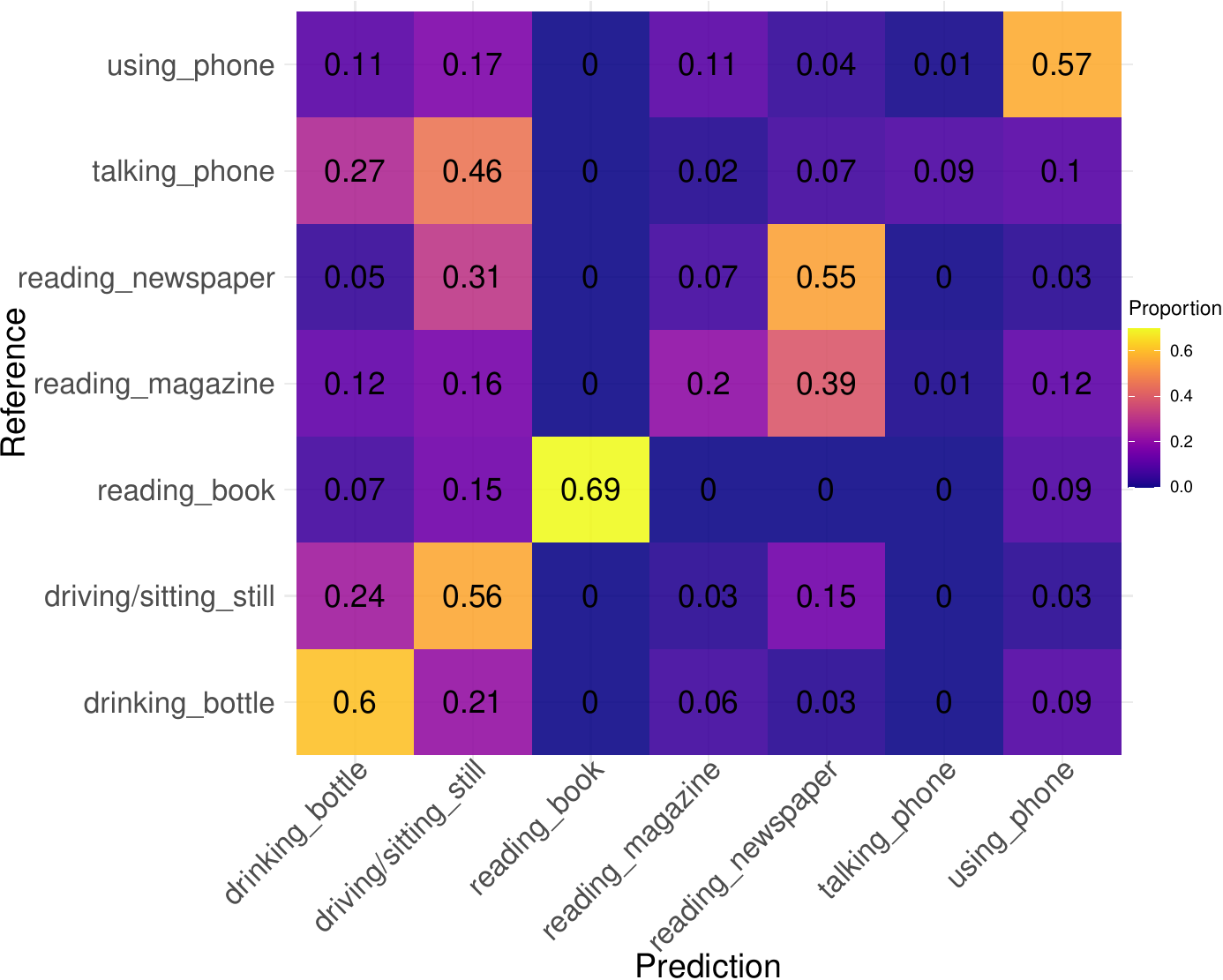}
    \caption{Confusion matrix for the direct  \textsc{Sim$\rightarrow$Real} transfer.}
    \label{fig:confmat}
\end{figure}

\begin{figure}
    \centering
    \includegraphics[width=0.5\textwidth]{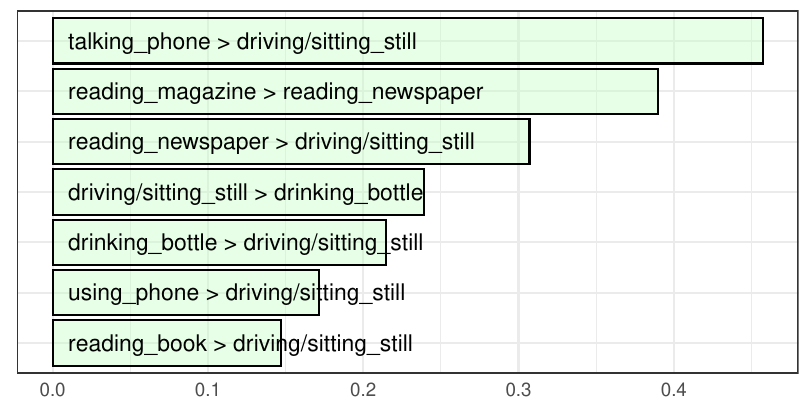}
    \caption{Most common confusions of an end-to-end CNN for driver observation after the direct  \textsc{Sim$\rightarrow$Real} transfer.}
    \label{fig:common_conf}
\end{figure}

Next, we analyze the most common confusions in Figure \ref{fig:common_conf} and the confusion matrix (Figure \ref{fig:confmat}).
Nearly all categories are often mistaken for \textit{driving/sitting still}, which is not surprising considering its overrepresentation in the original simulator-based training set~\cite{martin2019drive}. The category itself is recognized correctly in 56\% of the cases. It is unsurprising that \textit{reading a magazine} is most frequently confused with \textit{reading a newspaper} (39\%), while the confusion in the opposite direction is less common (7\%). \textit{Talking on the phone} is often mistaken for \textit{drinking} (27\%), which is understandable as both actions involve raising one hand close to the mouth. The difficulty in fine-grained recognition of smaller objects, such as phones, may arise from the downsampling performed by 3D CNNs, which rapidly decreases image resolution to obtain larger receptive fields.
Undoubtedly, \textit{talking on the phone} is a highly common and significant distractive secondary activity. Accurately recognizing such fine-grained actions from images holds immense importance for the future. Similarly, like other behaviors, the most frequent confusion for \textit{talking on the phone} is with the overrepresented category of \textit{driving/sitting still} (46\%). While \textit{using a phone} is better recognized than \textit{talking on the phone}, with a 57\% accuracy in predictions, confusions with \textit{driving/sitting still} still occur relatively frequently (17\%).

Considering this information, it is clear that improvements are needed, especially with regard to fine-grained activities involving smaller objects, particularly when differentiating activities from the ``default''  \textit{driving/sitting still} state.

\begin{figure*}
    \centering
    \includegraphics[width=\textwidth]{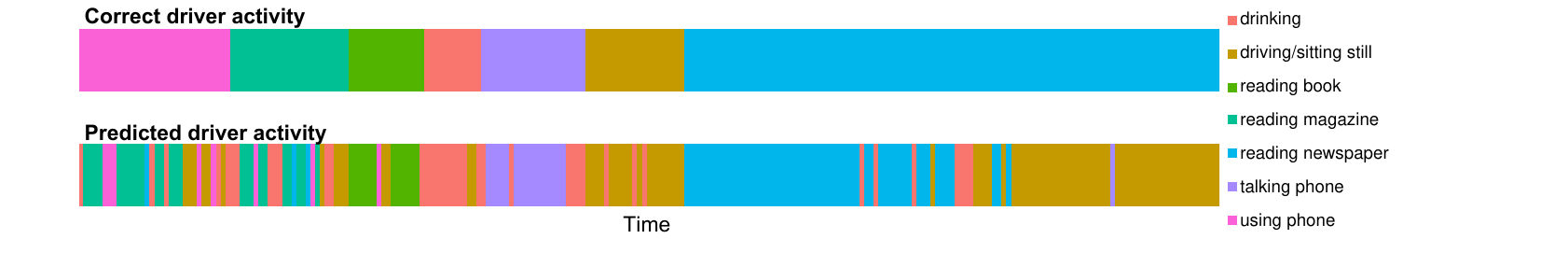}
    \caption{Example of a single driving session recorded in our dataset. The subject is engaged in six different secondary activities during fully autonomous driving. A brief manual driving segment is also present. The upper bar represents the ground truth activity labels, while the lower bar displays the predictions of an end-to-end  CNN trained on a simulator-based dataset. Different colors mark different secondary activities.
    }
    \label{fig:session}
\end{figure*}

\begin{table}[h!]
  \centering
  \caption{Recognition Quality for Driver Observation collected in a Simulator vs Real-life Self-Driving Car}
  \label{tab:accuracy}
  \begin{tabular}{ l c c }
    \toprule
    & \multicolumn{2}{c}{Accuracy} \\
    \multirow{-2}{*}{Activity} & Simulator & Real \\
    \specialrule{1pt}{0pt}{0pt}
    \rowcolor{gray!20}
    \multicolumn{3}{l}{\textbf{All Driver Activities}} \\
    \specialrule{1pt}{0pt}{0pt}
    Random chance baseline & 14.29 & 14.29 \\
    Simulator-trained I3D & 85.7 & 46.56 \\
    \specialrule{1pt}{0pt}{0pt}
    \rowcolor{gray!20}
    \multicolumn{3}{l}{\textbf{Per-category Results  for Simulator-trained I3D}} \\
    \specialrule{1pt}{0pt}{0pt}
    driving/sitting still & 99.01 & 55.76 \\
    using phone & 90.36 & 56.7 \\
    talking phone & 52.94 & 8.52 \\
    reading magazine & 73.91 & 20.34 \\
    reading newspaper & 98.03 & 54.72 \\
    reading book & - & 69.49 \\
    drinking & 100 & 60.4 \\
    \specialrule{1pt}{0pt}{0pt}
  \end{tabular}
\end{table}

\subsection{Attribution analysis and qualitative examples}

In Figure \ref{fig:session}, we visualize an example of a single driving session collected in our dataset. The upper bar depicts the true secondary activities of the driver, while the lower bar depicts the predictions of the I3D model trained on simulator data. The model had issues in recognizing the first two driver behaviors (\textit{using phone} and \textit{reading magazine}), although there were certain brief segments with the correct classification. Secondary behaviors that followed were easier to recognize, and the majority of the frames were assigned the correct label.
 A surprising observation is that, for this particular subject, \textit{talking on phone} was recognized better  than \textit{using phone}. This is contrary to the overall trend observed when examining the statistics of the entire dataset. These findings highlight that individual appearances and the unique manner in which humans perform actions can significantly influence the quality of recognition.

\begin{figure*}
  \centering

   \begin{subfigure}{0.24\linewidth}
    \includegraphics[width=\linewidth,  trim=0 1.5cm  0 0, clip]{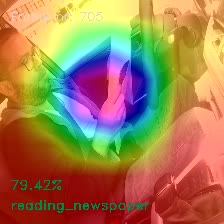}
    \caption{True: read newspaper \\ Pred: read newspaper \checkmark}
    \label{fig:subfig1}
  \end{subfigure}%
  \hfill
   \begin{subfigure}{0.24\linewidth}
    \includegraphics[width=\linewidth, trim=0 1.5cm  0 0, clip]{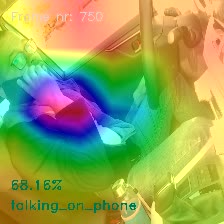}
    \caption{True: talk on phone \\ Pred: talk on phone  \checkmark}
    \label{fig:subfig2}
  \end{subfigure}%
  \hfill
  \begin{subfigure}{0.24\linewidth}
    \includegraphics[width=\linewidth, trim=0 1.5cm  0 0, clip]{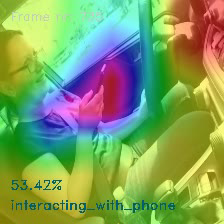}
    \caption{True: interact with phone \\ Pred: interact with phone   \checkmark}
    \label{fig:subfig3}
  \end{subfigure}%
  \hfill
  \begin{subfigure}{0.24\linewidth}
    \includegraphics[width=\linewidth, trim=0 1.5cm  0 0, clip]{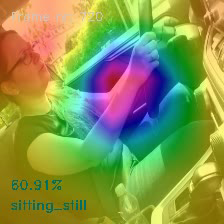}
    \caption{True: sitting still / driving  \ Pred: sitting still / driving \checkmark}
    \label{fig:subfig4}
  \end{subfigure}

  \vspace{0.5cm}

  \begin{subfigure}{0.24\linewidth}
    \includegraphics[width=\linewidth, trim=0 1.5cm  0 0, clip]{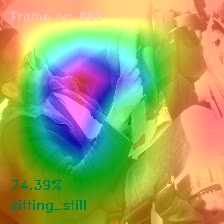}
    \caption{True: read newspaper \\ Pred: sitting still / driving \ding{55}}
    \label{fig:subfig5}
  \end{subfigure}%
  \hfill
  \begin{subfigure}{0.24\linewidth}
    \includegraphics[width=\linewidth, trim=0 1.5cm  0 0, clip]{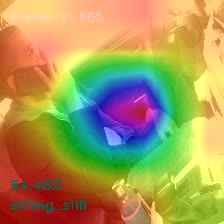}
     \caption{True: talk on phone \\ Pred: sitting still / driving \ding{55}}
    \label{fig:subfig6}
  \end{subfigure}
  \hfill
  \begin{subfigure}{0.24\linewidth}
    \includegraphics[width=\linewidth, trim=0 1.5cm  0 0, clip]{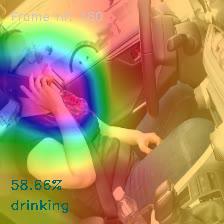}
     \caption{True: talk on phone \\ Pred: drinking \ding{55}}
    \label{fig:subfig7}
  \end{subfigure}
  \hfill
  \begin{subfigure}{0.24\linewidth}
    \includegraphics[width=\linewidth, trim=0 1.5cm  0 0, clip]{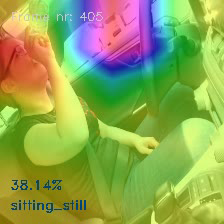}
     \caption{True: drinking \\ Pred: sitting still / driving \ding{55}}
    \label{fig:subfig8}
  \end{subfigure}

  \caption{Qualitative results and  attribution analysis: image portions with high activation values in the intermediate network layers are highlighted as a heatmap. These regions are estimated via a spatiotemporal implementation of GradCAM~\cite{selvaraju2016grad}. 
  }
  \label{fig:gradcam}
\end{figure*}

In our subsequent analysis, we explore the specific input pixels that influence the model's decisions, leveraging the GradCAM technique~\cite{selvaraju2016grad} adapted for our spatiotemporal data (see Section \ref{sec:gradcam}).
Figure \ref{fig:gradcam} presents frames that have been correctly classified (top row) and those misclassified (bottom row), with an overlay of a heatmap that highlights the pixels that contributed  to the current network decision.
The analysis unveils that the video classification model predominantly focuses on the hands and  objects in use, particularly when the predictions are accurate. Interestingly, in the correctly classified \textit{talk on phone}  example (Figure \ref{fig:subfig2}), the attention does not center on the hands, but gravitates towards the area around the wrist. This observation suggests that the network places importance not on the object per se, but rather on a specific wrist position, potentially accounting for the subpar recognition of this category.
Conversely, when \textit{talk on phone}is erroneously identified as \textit{drinking} (Figure \ref{fig:subfig7}), the model’s attention is indeed trained on the driver's hands. This evidence suggests that the model's comprehension of \textit{talk on phone} and \textit{drinking} is centered around certain wrist and hand patterns, discounting the significance of the associated objects.
We also observe, especially in failure cases, that the model's attention can deviate from the hands or the relevant object, steering towards the mid-cabin area or certain outside patterns, presumably in response to unusual movements absent from the simulator training data.
In conclusion, the GradCAM analysis offers significant insights into the way simulator-based video classification models operate when facing naturalistic driving data.  Especially for categories that are hard to recognize, we discover a strong reliance of the model on hand and wrist movements, potentially at the expense of object recognition. This might be due to the different appearance of these objects present in the training set. 
Enhancing the training data with a broader range of phones, drinking bottles, cups, and similar objects could address this limitation. Furthermore, we observe instances where the model's focus shifts to external movement during misclassifications (Figure \ref{fig:subfig8}). This suggests that the absence of such movement in the simulator data adversely impacts the model's recognition capability.
Moving forward, we advocate for the inclusion of such variable movements in the training data. This could be accomplished by recording more naturalistic datasets or developing sophisticated data augmentation methods that effectively mimic these car movements.

\section{Conclusion}

We collected a video-based dataset for driver activity recognition during real autonomous driving sessions. Our key motivation is to study the  direct \textsc{Sim$\rightarrow$Real} transfer of deep learning-based driver observation models, which is of particular relevance given that simulated data is a prevalent resource in autonomous driving research.
Our dataset features seven drivers engaged in six distractive  activities as well as a short manual driving segment.  Furthermore, the dataset is constructed with annotations and sensor correspondence to a large-scale simulator-based dataset, specifically designed to supplement the validation protocol of simulator-trained models with real-world data.
While the  model clearly surpasses the random baseline, its recognition quality drops drastically when moving from simulated to real-life recordings, highlighting  the necessity of incorporating real-world data into validation protocols of simulator-trained models.

\mypar{Acknowledgements.} This work was supported by the Austrian Ministry for Climate Action, Environment, Energy, Mobility, Innovation, and Technology (BMK) Endowed
Professorship for Sustainable Transport Logistics 4.0., IAV France S.A.S.U., IAV GmbH, Austrian Post AG and the UAS
Technikum Wien. Alina Roitberg was partially supported by the KHYS Connecting Young Scientists travel award and Deutsche Forschungsgemeinschaft (DFG) under Germany’s Excellence Strategy - EXC 2075.

\balance
\bibliographystyle{ieeetr}
\bibliography{bib}

\end{document}